%% file: anonymous-submission-latex-2026.tex
\documentclass[letterpaper]{article} 
\usepackage{aaai2026}  
\usepackage{times}  
\usepackage{helvet}  
\usepackage{courier}  
\usepackage[hyphens]{url}  
\usepackage{graphicx} 
\usepackage{natbib}  
\usepackage{caption} 
\usepackage{algorithm}
\usepackage{algorithmic}

\usepackage{newfloat}
\usepackage{listings}

\usepackage{multirow}
\usepackage{cite}
\usepackage{amsmath}
\usepackage{booktabs}
\usepackage{amsfonts}

\DeclareCaptionStyle{ruled}{labelfont=normalfont,labelsep=colon,strut=off} 
\floatstyle{ruled}
\newfloat{listing}{tb}{lst}{}
\floatname{listing}{Listing}
\title{LOG-Nav: Efficient Layout-Aware Object-Goal Navigation with \\Hierarchical Planning}
\author{
    Jiawei Hou\textsuperscript{\rm 1},
    Yuting Xiao\textsuperscript{\rm 2},
    Xiangyang Xue\textsuperscript{\rm 1,}$^*$,
    Taiping Zeng\textsuperscript{\rm 2,}\thanks{Co-corresponding author}
}
\affiliations{
    \textsuperscript{\rm 1}School of Computer Science, Fudan University, Shanghai, China\\
    \textsuperscript{\rm 2}Institute of Science and Technology for Brain-Inspired Intelligence, Fudan University, Shanghai, China

    \{jwhou23, ytxiao25\}@m.fudan.edu.cn, 
    \{xyxue, zengtaiping\}@fudan.edu.cn
}

\usepackage{bibentry}

\begin{document}

\maketitle

\begin{abstract}
We introduce LOG-Nav, an efficient layout-aware object-goal navigation approach designed for complex multi-room indoor environments. 
By planning hierarchically leveraging a global topologigal map with layout information and local imperative approach with detailed scene representation memory, LOG-Nav achieves both efficient and effective navigation.
The process is managed by an LLM-powered agent, ensuring seamless effective planning and navigation, without the need for human interaction, complex rewards, or costly training.
Our experimental results on the MP3D benchmark achieves 85\% object navigation success rate (SR) and 79\% success rate weighted by path length (SPL) (over 40\% point improvement in SR and 60\% improvement in SPL compared to exsisting methods). Furthermore, we validate the robustness of our approach through virtual agent and real-world robot deployment, showcasing its capability in practical scenarios.
\end{abstract}

\begin{links}
    \link{Code}{https://github.com/fudan-birlab/LOGNav}
\end{links}

\input{sections/introduction}
\input{sections/related}
\input{sections/overview}
\input{sections/method}
\input{sections/experiments}
\input{sections/limitations}


\bibliography{aaai2026}


\end{document}

%% file: sections/introduction.tex
\section{Introduction}\label{introduction}

Advancements in Large Foundation Models (LFMs) and robotics research have brought household assistant robots closer to real-world deployment. A key capability for such robots is to locate a target and navigate to it based on user input~\citep{zhu2017target}. Researchers have made significant progress in integrating Vision-Language Models (VLMs) to align user instructions with robot observations for prompted navigation and target localization using scene memory collected during exploration~\citep{shah2023lm, zhou2023esc, dorbala2022clip, chen2019touchdown, vasudevan2021talk2nav, krantz2023iterative}. 
While complementary scene understanding and accurate instruction interpretation are essential for robot navigation, two mainstream approaches have emerged: process-prompted and goal-oriented navigation.

Recent advancements in robot navigation have focused on understanding and executing user instructions by leveraging Vision-Language Models (VLMs). These models enable robots to follow process-prompted instructions, allowing them to explore and navigate in unseen environments~\citep{anderson2018vln, chen2019touchdown}.
However, these benchmarks are based on sparse panoramic observations, assuming known topologies, oracle navigation, and perfect localization, which do not reflect real-world deployment challenges~\citep{vlnce}. 
VLN-CE has attempted to bridge this gap by removing unrealistic assumptions, translating observations into low-level controls in continuous environments~\citep{vlnce}.
The grounding of process-prompted instructions is in a step-by-step form, such as ``Walk out of the bedroom. Turn right and walk down the hallway. At the end of the hallway turn left.''. 

\begin{figure}[t]
\centering
\includegraphics[width=\linewidth]{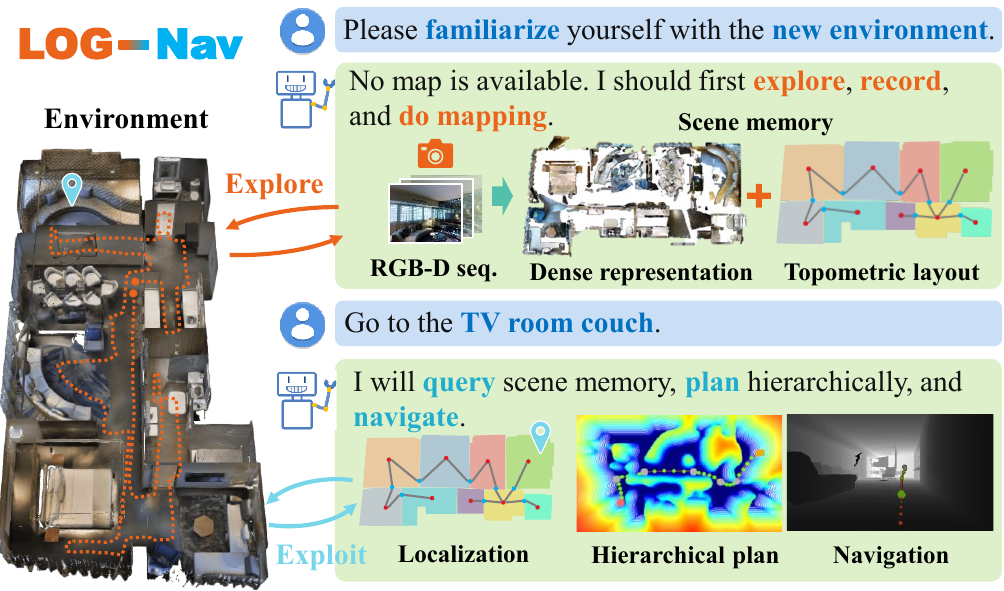}
\caption{\textbf{Our proposed LOG-Nav.} An efficient object-goal navigation approach with LLM-powered agent that realizes hierarchical planning based on topology and dense scene memory. The entire process is conducted automatically without costly training or complex rewards.}
\label{fig::teaser}
\end{figure}

\begin{figure*}[t]
\centering
\includegraphics[width=0.9\linewidth]{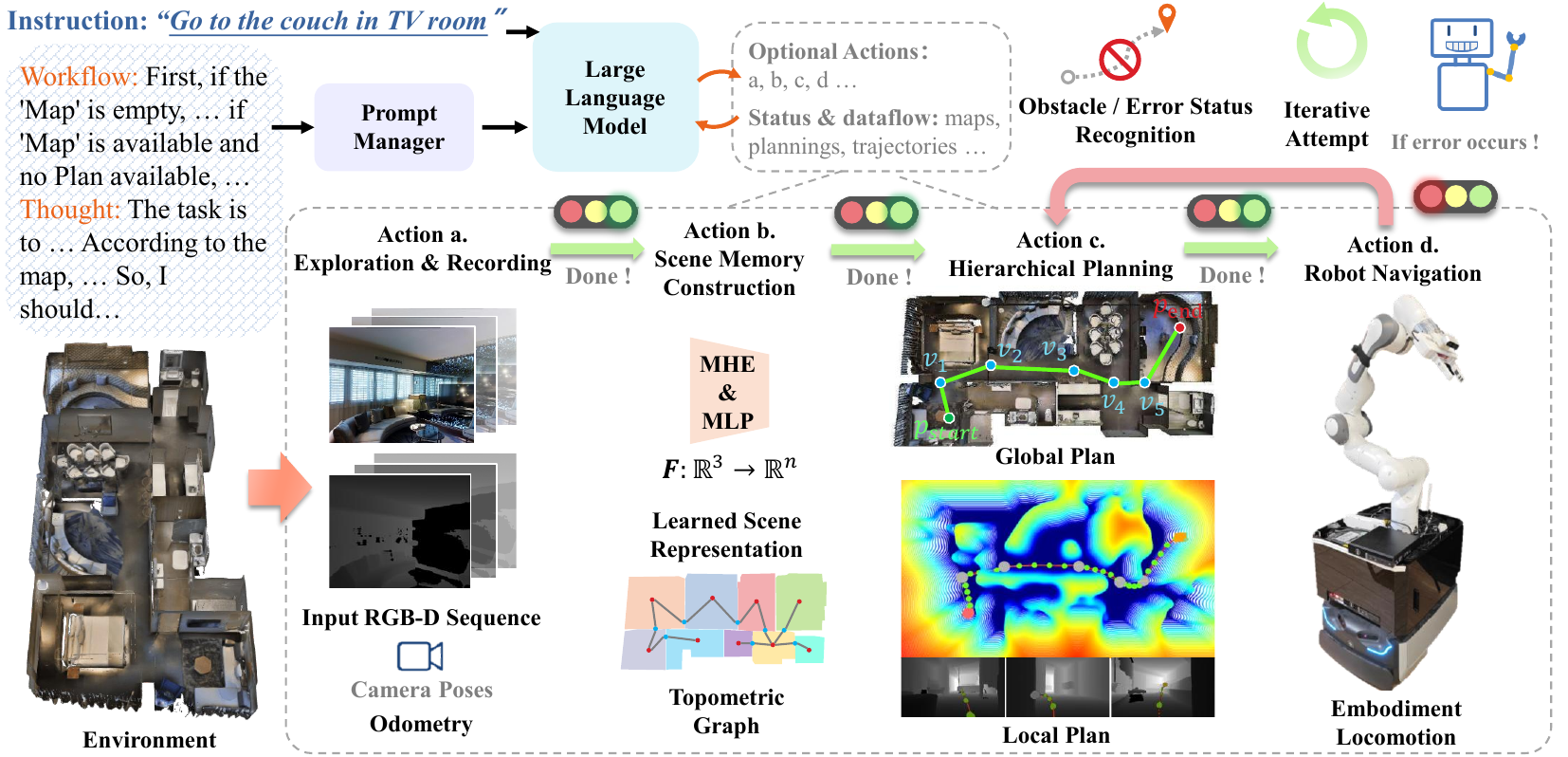}
\caption{\textbf{Overview of our proposed method.} The LLM agent takes user instructions as input and manages the optional action choices according to the prompts and data flow. Optional actions include exploring and recording the scene, constructing scene memory representation, planning, and executing navigation. Obstacles, error recognition, and iterative attempts are available.}
\label{fig::method}
\end{figure*}

While following step-by-step instructions is a necessary skill for household assistants, the goal-oriented task provides a simple optional interaction pattern for users. For instance, robots should interpret simple instructions like ``Go to the couch in the living room'' or a picture of the desired couch. Since ZSON~\citep{majumdar2022zson} proposed the ability to locate and navigate to the object of interest in a scene based on a user inquiry, without training on this specific setting, researchers have made significant efforts to improve retrieval processes, instance recognition abilities, and scene understanding~\citep{guan2024loczson, psl, zhou2023esc, ramakrishnan2022poni, yu2023l3mvn}.
Despite these advancements, navigation in open scenes remains a challenging problem due to practical constraints. Specifically, the following challenges persist: 1) While targets may not be directly observed, planning globally efficient paths based on scene memory without unnecessary detours is important. 2) Adapting to local scene changes that may occur compared to the initial scene memory. 
3) Realizing effective planning without human aid, complex rewards, or costly training.

Recent works in robot planning have demonstrated diverse approaches. Works like L3MVN~\citep{yu2023l3mvn}, ESC~\citep{zhou2023esc}, and VLFM~\citep{yokoyama2024vlfm} employ LFMs to respectively map the scene with language-guided frontier exploration, probabilistic commonsense constraints, and visual-language similarity scoring, bridging semantic understanding with geometric planning. Taking global layouts into consideration, HOVSG~\citep{werby23hovsg} constructs open-vocabulary 3D scene topological graphs through feature clustering to enable navigation.
However, the static topological way-points extracted from occupancy cannot ensure efficient path planning or manage unexpected local scene changes.
To solve this dual-challenge, we propose leveraging a hierarchical framework uniquely integrates a lightweight global topological map for coarse route planning with a local imperative approach that dynamically adjusts trajectories. 

In this work, we introduce LOG-Nav, an efficient layout-aware object-goal navigation framework designed for complex multi-room indoor environments. 
LOG-Nav leverages both topological layout relationships and detailed scene representations to achieve efficient navigation in a hierarchical framework. By employing the dual-level information, the planning process first queries the open-world semantics based on user instructions to generate a global navigable topometric path at the layout level.
The global waypoints are then sequentially projected into the robot's ego-centric observation space, enabling the robot to plan a dense local waypoint set that accounts for scenario changes. 
An LLM-powered agent is employed that operates without human interaction, complex reward design, or costly training on specific settings.
The method requires only RGB-D observations as input and outputs navigation results as 3D waypoint sets, which are compatible with general platforms. The primary functions are shown in Fig.\ref{fig::teaser}. Evaluated on the widely-used MP3D dataset~\citep{ramakrishnan2021hm3d}, LOG-Nav demonstrates superior navigation success rates (SR) and efficiency, as measured by success weighted by path length (SPL), compared to exisiting methods. 
Additionally, we validate our framework through both virtual agent and real-world robots, demonstrating its applicability, efficiency and robustness against unexpected scene changes.

Our contributions are summarized as follows:

\begin{itemize}
    \item We present LOG-Nav, an efficient layout-aware object-goal navigation approach that leverages hierarchical planning, managed by an LLM-powered agent without further rewards or training.
    \item We develop a novel hierarchical strategy that integrates global topological layout-aware planning with a local dynamic-aware imperative approach, ensuring efficient and effective navigation both globally and locally in complex indoor environments.
    \item Complex experiments are conducted on the MP3D dataset demonstrating a 85\% SR and 79\% SPL in object navigation. Deployments on both virtual agent and real robots are provided to verify the applicability.
\end{itemize}

%% file: sections/related.tex
\section{Related Works}

\begin{figure}[t]
\centering
\includegraphics[width=\linewidth]{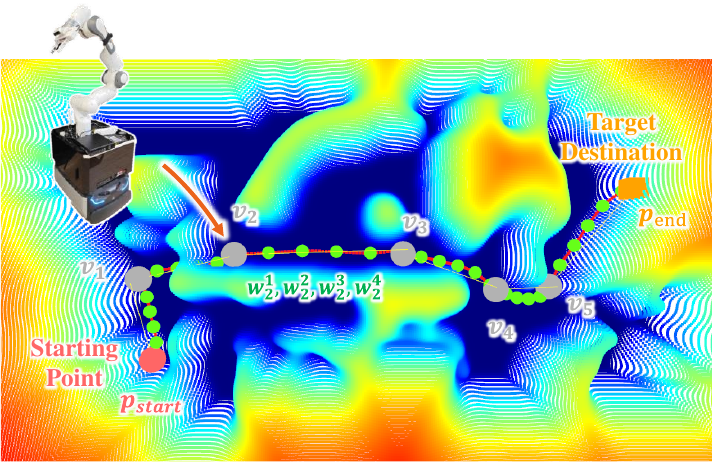}
\caption{\textbf{Hierarchical planning example.} Global planning is conducted on the topological graph and generates $V=\{v_1, v_2, \dots , v_n\}$. Local planning, realized in IL approach, generates $W_{i,i+1}=\{w_i^1, w_i^2, \dots , w_i^m\}$.}
\label{fig::plan}
\end{figure}

\subsection{Process-prompted Navigation}

Vision-Language navigation requires agents to follow free-form linguistic instructions to navigate in unseen environments~\citep{anderson2018vln}, necessitating an understanding of complex observations and the ability to interpret instructions for effective navigation. Pioneering works were conducted under the simplified assumption that environments could be abstracted as discrete connectivity graphs~\citep{anderson2018vln, vlnce}. To bridge the gap for more realistic modeling, researchers have dedicated efforts to address the limitations imposed by task settings~\citep{vlnce, zhang2024navid, krantz2021waypoint, xu2023vision, anderson2021sim}. However, the assumption of closed-set scene priors limits the agent to pre-mapped environments. Additionally, VLN evaluation metrics often emphasize strict path fidelity over goal discovery efficiency. On the contrary, we follow a goal-oriented paradigm and propose open-set scene-specific representations built during exploration and object navigation, executed in a human-interaction-free manner with LLM agent.

\subsection{Goal-oriented Navigation}
Goal-oriented navigation involves interpreting multi-modal target descriptions, localizing, and navigating to the specified targets. Notable works~\citep{yang2018visual, du2020learning, campari2020exploiting, liang2021sscnav, chaplot2020object} have demonstrated impressive capabilities using large-scale reinforcement learning. However, these methods typically require substantial training data and computational resources and may suffer when generalizing across diverse environments and embodiments~\citep{ramakrishnan2022poni}. ZSON~\citep{majumdar2022zson} trains navigation agents in target environments by mapping object goals to image-goal embeddings, emphasizing the ability to locate and navigate to objects based on inquiries without specific training in that setting. Further advancements have been made~\citep{guan2024loczson, psl} to enhance image retrieval and semantic understanding, achieving higher success rates. Additionally, L3MVN~\citep{yu2023l3mvn} has improved exploration efficiency by constructing environment maps using frontier-based methods and leveraging LLM inference. ESC~\citep{zhou2023esc} introduced soft commonsense constraints through probabilistic logic, enabling training-free zero-shot navigation by embedding object-room relationships, while VLFM~\citep{yokoyama2024vlfm} eliminated text dependency by directly grounding visual features with language models through RGB-based similarity scoring, significantly accelerating semantic processing. Nonetheless, the absence of explicit global layout-level guidance limits their ability to ensure efficient path planning. Works like HOVSG~\citep{werby23hovsg} construct open-vocabulary 3D scene topological graphs through feature clustering to enable navigation. 
However, the static topological way-points extracted from occupancy cannot ensure efficient path planning or manage unexpected local scene changes.

%% file: sections/overview.tex
\section{Overview}
\label{overview}

We propose a hierarchical robot navigation framework for operation in unexplored indoor environments, guided by natural language instructions $T$, images $I$, or 3D locations $P$. The framework overview is shown in Fig.\ref{fig::method}. 
The planning process is conducted on a dual-level map with global topology and local dense representation built with collected RGB-D sequences ${i, d}$, where $i$ and $d$ denote RGB and depth observations, respectively.
The scene representation consists of (1) an implicit neural function $F:\mathbb{R}^3 \to \mathbb{R}^n$ maps 3D positions to vision-language embeddings; and (2) a topometric map $G=(V, E)$, where $V$ comprises region vertices $\mathbf{v}_r$ and entrance vertices connecting regions $\mathbf{v}_e$ and edges $E$ connecting these vertices.

During planning, inputs $T/I$ are encoded into embeddings $\mathbf{E}$ using a pre-trained vision-language model. These embeddings are then queried in $F$ to identify the target position $p_{\text{end}}$ with the highest matching similarity. A global navigable path is computed between the robot's current pose $p_{\text{start}}$ and $p_{\text{end}}$ using the topometric map $G$, generating a sparse sequence of waypoint vertices $V=\{v_1, v_2, \dots , v_n\}$. 
These waypoints are progressively refined during navigation in an imperative approach, with $W_{i,i+1}=\{w_i^1, w_i^2, \dots , w_i^m\}$ denoting the densified local waypoints between $v_i$ and $v_{i+1}$. Navigation proceeds incrementally: as the robot approaches $v_{i+1}$, the next segment $W_{i+1, i+2}$ is planned.
The framework incorporates an LLM agent designed to integrate essential data streams for task execution.

%% file: sections/method.tex
\section{Method}\label{method}

\subsection{Hierarchical Scene Representation}

Recent works such as HOV-SG~\citep{werby23hovsg}, CLIO~\citep{Maggio2024Clio}, and Topo-Field~\citep{hou2024topofield} have demonstrated progress in representing scenes as topo-maps while retaining detailed content. Although topometric maps offer computational efficiency for downstream planning and navigation tasks, high-fidelity scene content from neural implicit representations enables semantic cue queries~\citep{hou2024topofield}. Our approach represents 3D scenes combining a high-level topometric map and a detailed neural implicit representation.

Given posed RGB-D sequences $\{i, d\}$, we train a neural implicit function $F$~\citep{hou2024topofield} to match 3D points $p$ derived from depth and camera parameters with the encoded vision-language embedding $\mathcal{E}_p$~\citep{clip, sentence-bert}. A topometric map $G = (V, E)$ is then extracted with vertices $V$ including region vertices $\mathbf{v}_r$ and entrance vertices $\mathbf{v}_e$ and edges $E$ connecting them. The subsequent scene representation is denoted as:
\begin{align}
    F&:\mathbb{R}^3 \to \mathbb{R}^n, \\
    \notag
    G&=(V, E).
\end{align}

\subsection{Hierarchical Planning}\label{planning}

Our planning design strategy encompasses two complementary objectives: 1) The planned path must be globally efficient, avoiding unnecessary deviations and minimizing twists and turns. In indoor scenes, given that the target destination may not be directly observable, robots should avoid entering irrelevant rooms or revisiting explored areas. 2) The robot must possess the capability to detect and adapt to minor changes in the environment, such as the appearance of unexpected obstacles in spaces previously identified as free.

\subsubsection{Global Topology Plan}

To ensure globally efficient path planning for inter-room navigation, our approach utilizes a topological graph to generate a sparse waypoint set that the robot must traverse to reach the destination, typically located at room entrances or connecting doors. The process begins by unifying multi-modal instructions (images, text, or 3D positions) into a destination 3D position. The text/image instruction $T/I$ is first encoded using CLIP~\citep{clip} and Sentence-BERT~\citep{sentence-bert} to generate the target embedding $\{C, S\}_{T/I}$. A point set $P^*$ is then sampled uniformly from the environment, and $F(P^*)=\{(f_v,f_s)\}_{P^*}$, representing the vision and semantic embeddings of the corresponding 3D positions, is calculated. The cosine similarity between $\{C, S\}_{T/I}$ and $F(P^*)$ determines the position $p_{\text{end}}\in P^*$ with the highest similarity to the instruction query~\citep{hou2024topofield}. The topological graph $G$ is subsequently queried using the robot's current position $p_{\text{start}}$ and $p_{\text{end}}$ to identify the shortest path connecting them. This results in a sequence of vertices $V=\{v_1, v_2, \dots , v_n\}$, where each $v_i$ represents a structural element, such as doors or corridors, that the robot must navigate through to reach the destination.

\subsubsection{Local Ego-centric Plan}

In the initial global planning phase, a sparse set of waypoints denoted as $V=\{v_1, v_2, \dots , v_n\}$, is established. These waypoints guide the robot's inter-room navigation, ensuring efficient travel between distinct regions. To achieve intra-room movement and local navigation precision, dense waypoints, $W_{i,i+1}=\{w_i^1, w_i^2, \dots , w_i^m\}$, are computed between consecutive global waypoints $v_i$ and $v_{i+1}$. We realize local planning in an Imperative Learning (IL) approach with the target way-point $v_{i+1}$ in the global coordinates and current depth observation $d'$ with pose $p'$. First, $v_{i+1}$ is projected to the depth frame with the pose $p'$. The depth input, encoded by a convolutional network~\citep{lecun1995convolutional} into a depth embedding $\mathcal{O_{d'}}$, together with the way-point feature $\mathcal{P'}$ of $p'$ is input into a planning network~\citep{yang2023iplanner} that generates a set of local key points $W_{i,i+1}$ from
\begin{equation}
    \tau = \mathcal{F}(\mathcal{O}_{d'}, \mathcal{P'}),
\end{equation}
where $\mathcal{F}$ is the planner and $W_{i,i+1}$ is the discrete point set sampled from the planned trajectory $\tau$.
As the robot approaches each subsequent global waypoint $v_{i+1}$, the next segment of dense waypoints $W_{i+1, i+2}$ is computed and executed incrementally.

The hierarchical pathway from the robot's starting position $p_{start}$ to the destination $p_{end}$ is structured as
\begin{align}
    p_{\text{start}}\to p_{\text{end}}&: V=\{v_1, v_2, \dots , v_n\}, \\
    \notag
    v_i \to v_{i+1}&: W_{i,i+1}=\{w_i^1, w_i^2, \dots , w_i^m\}.
\end{align}

This two-tiered approach ensures efficient navigation at both the global and local levels, allowing the robot to adapt dynamically to changes in the environment while maintaining a clear path. An illustration is shown in Fig.\ref{fig::plan}.

\subsection{LLM-powered Navigation Agent}

\begin{figure}[t]
\centering
\includegraphics[width=\linewidth]{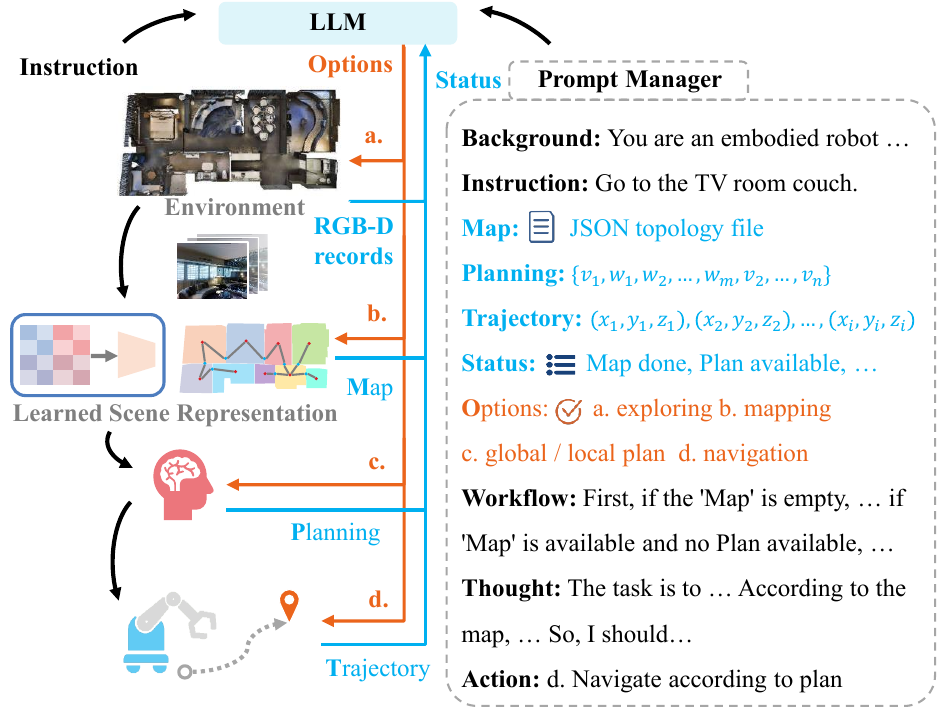}
\caption{\textbf{Data flow example of LLM Agent.} Instructed by the user input, the agent conducts prompts, applies optional actions, and gets feedback from interaction.}
\label{fig::agent}
\end{figure}

\begin{table*}[t]
\centering
\setlength{\tabcolsep}{2pt}
\begin{tabular}{lcccccccc}
\toprule 
\multirow{2}{*}{\textbf{Methods}} & \textbf{Object} &\textbf{Nav.} &  & \textbf{Instance } & \textbf{Nav. (text-goal)} & & \textbf{Instance } & \textbf{Nav. (image-goal)} \\ \cline{2-3} \cline{5-6} \cline{8-9}
                      & SR & SPL & & SR & SPL & & SR & SPL \\ 
\midrule
CoW (2022)~\citep{gadre2022cow}                 & 0.061 & 0.039 & & 0.018 & 0.011 & & - & - \\
ZSON (2022)~\citep{majumdar2022zson}                & 0.255 & 0.126 & & 0.106 & 0.049 & & 0.146 & 0.073 \\
L3MVN (2023)~\citep{yu2023l3mvn}               & 0.352 & 0.165 & & - & - & & - & - \\
ESC (2023)~\citep{zhou2023esc}                 & 0.392 & 0.223 & & 0.065 & 0.037 & & - & - \\
VLFM (2024)~\citep{yokoyama2024vlfm}                         & 0.364 & 0.175 & & - & - & & - & - \\
PixNav (2024)~\citep{cai2024pixnav}              & 0.379 & 0.205 & & - & - & & - & - \\
HOV-SG (2024)~\citep{werby23hovsg}     & 0.404 & 0.236 & & - & - & & - & - \\
PSL (2024)~\citep{psl}                 & 0.424 & 0.192 & & 0.165 & 0.075 & & 0.230 & 0.114 \\
LOG-Nav(Ours)     & \textbf{0.856} & \textbf{0.797} & & \textbf{0.786} & \textbf{0.701} & & \textbf{0.673} & \textbf{0.582} \\
\bottomrule 
\end{tabular}
\caption{\textbf{Quantitative comparison of object navigation and instance navigation tasks on MP3D dataset.} For object navigation, text instruction is used as input. For instance navigation, text-prompted and image-prompted instructions are separately validated. The SR and SPL are employed as metrics.}
\label{exp_sim}
\end{table*}

\begin{table*}[t]
\small
\centering
\setlength{\tabcolsep}{4.2pt}
\begin{tabular}{lcccccc}
\toprule 
\textbf{Targets}             & Obstacles & Local-Plan Retries & Global-Plan Retries & Path Length (m) & Time Cost (s) & \textbf{SR} \\ 
\midrule
chair (text)        & - & 0/10 & 0/10 & 16 & 91 & 10/10 \\
chair (image)       & - & 2/10 & 0/10 & 16 & 93 & 7/10 \\
sink  (text)        & 1 & 2/10 & 0/10 & 20 & 94 & 9/10 \\
sink  (image)       & 1 & 3/10 & 1/10 & 20 & 97 & 6/10 \\
sofa  (text)        & 2 & 3/10 & 2/10 & 15 & 82 & 6/10 \\
sofa  (image)       & 3 & 6/10 & 3/10 & 15 & 80 & 3/10 \\
\bottomrule 
\end{tabular}
\caption{\textbf{Quantitative object navigation evaluation in real-world scene.} Each navigation task is evaluated 10 times from a similar starting position to the same target. \textit{Obstacles} means the number of obstacles we set after robot exploration where robots have to bypass to approach targets. \textit{Local/Global-Plan Retries} means the number of replanning called by the agent during the 10 times of navigation. \textit{Path Length} is the average path length robot travels in 10 times navigation. \textit{Time cost} is the average cost from starting points to destinations, where failure cases are not counted. \textit{SR} is the navigation success rate.}
\label{exp_real}
\end{table*}

Intuitively, a household robot equipped with intelligent algorithms and LLMs-powered knowledge should automatically conduct the necessary process to get into work in a new environment. To address this, an advanced LLM-powered agent is employed to manage the autonomous operation of a household robot in unexplored environments which oversees both exploration and exploitation phases.
The agent's design is guided by two overarching principles: 1) It should integrate user task descriptions, representations of the scene, and the robot's current state to make informed decisions; 2) It employs a comprehensive set of strategies, including mapping, path planning, and locomotion to effectively accomplish tasks while maintaining the ability to self-update. At each decision-making step, the agent processes a sequence of contextual tokens, denoted as:
\begin{equation}
    \{\mathbf{B}, \mathbf{I}, \mathbf{M}, \mathbf{P}, \mathbf{T}, \mathbf{S}, \mathbf{O}, \mathbf{A}\}
\end{equation}
where $\mathbf{B}$ represents the task background information, structured as ``You are an embodied robot ...''. $\mathbf{I}$ denotes the user's instruction, which can be text, an image, or a 3D position. $\mathbf{M}$ is the hierarchical map of the environment, comprising a learned implicit function $F$ and a topological graph $G$ in JSON format. $\mathbf{P}$ includes the robot's historical planning data. $\mathbf{T}$ refers to the robot travel trajectory. $\mathbf{S}$ indicates the robot's current status, such as its location and whether the planned path remains viable. $\mathbf{O}$ represents any optional skills the robot may use. $\mathbf{A}$ is the action decision.

The primary functions of the agent are as follows:
\begin{itemize}
    \item Mapping. The robot captures RGB-D images to construct a comprehensive environmental representation using neural implicit functions and topometric maps.
    \item Planning. Based on the scene representation $(F, G)$ and user instructions, the agent conducts planning described in Section \ref{planning} to fulfill the task.
    \item Navigation. The robot follows the planned waypoints to execute the designated actions and recognize obstacles or errors to conduct iterative attempts.
\end{itemize}

The LLM directs the operational workflow by selecting the most suitable actions, primarily informed by above mentioned information. Upon entering a new environment, the robot initiates a mapping process, exploring the space frontier-based to gather RGB-D observations and build its scene representation. Once the scene is understood, the agent proceeds task planning. Continuous monitoring trajectory and status ensures that navigation proceeds as intended. If errors arise, such as repeated visits, excessive time expenditure without reaching the next waypoint, or exceeding re-planning attempts, the LLM either replans or issues an error report. Fig.\ref{fig::agent} shows entire process and primary data flow.

%% file: sections/experiments.tex
\section{Experimental Results}
\label{experiment}

\begin{figure*}[t]
\centering
\includegraphics[width=\linewidth]{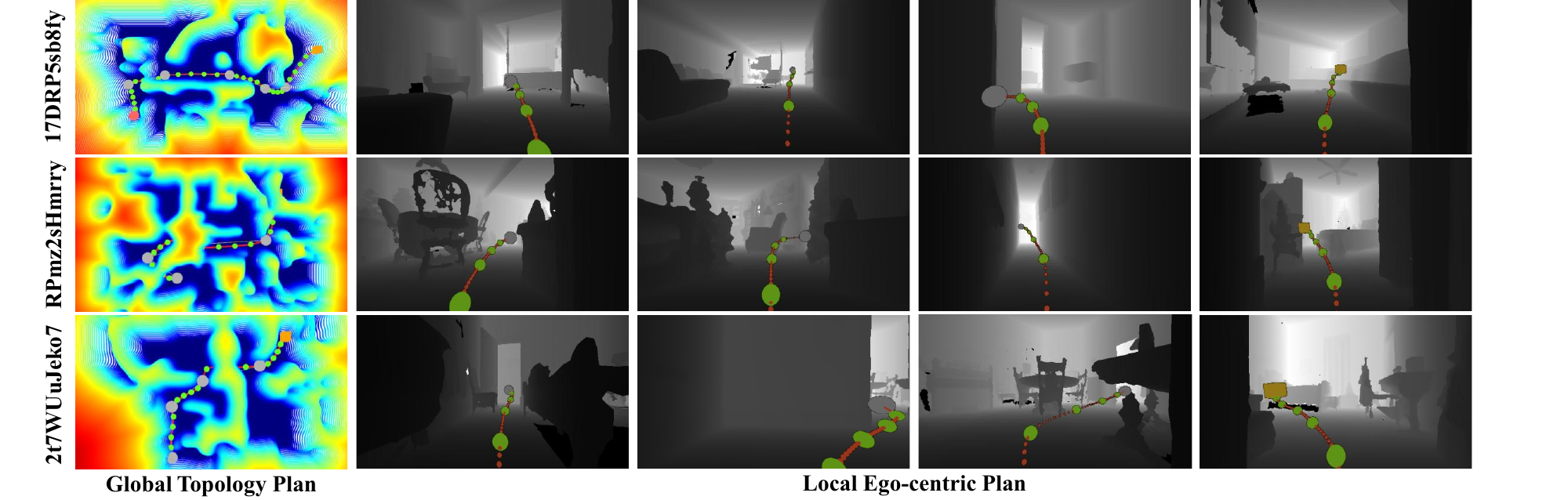}
\caption{\textbf{Navigation results on MP3D dataset.} The left column shows global planning results, where gray points are entrance vertices in the topological map, red points show the starting positions and yellow points show the destinations. The right columns show local planning results on the ego-centric views.}
\label{fig::sim_experiment}
\end{figure*}

\begin{figure}[t]
\centering
\includegraphics[width=\linewidth]{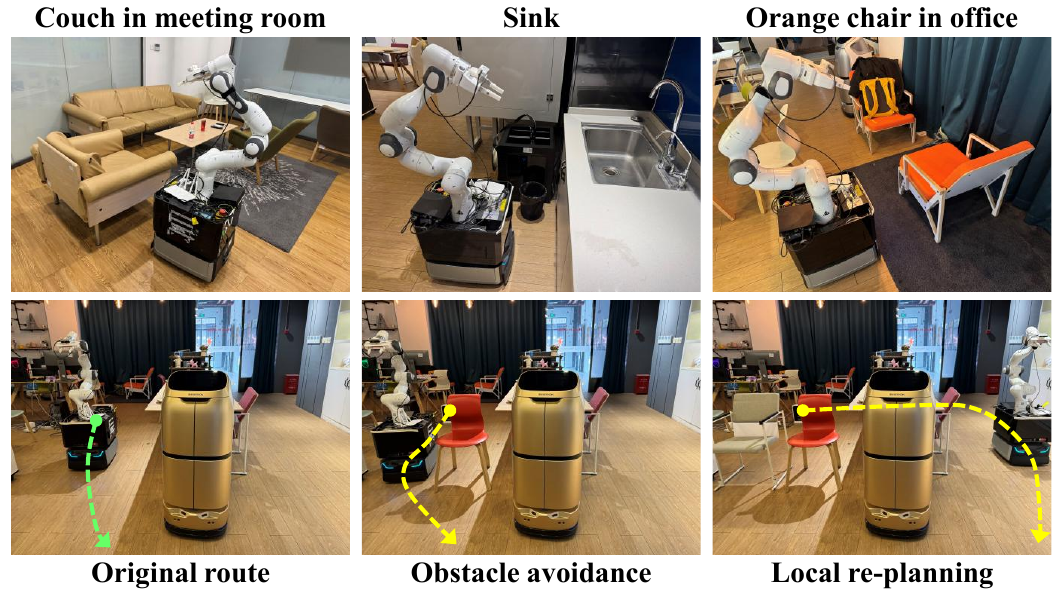}
\caption{\textbf{Real-world experiment results with mobile robots.} The above row shows the navigation success examples. The bottom row shows the unexpected obstacle avoidance examples compared to the built scene memory.}
\label{fig::real_experiment}
\end{figure}

\begin{figure}[t]
\centering
\includegraphics[width=\linewidth]{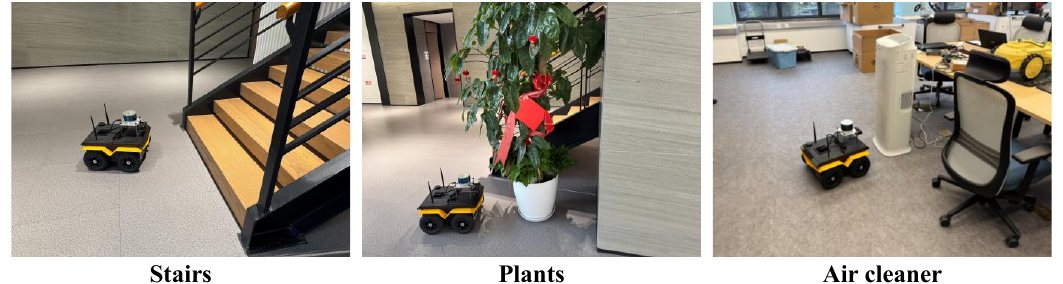}
\caption{\textbf{Real-world experiment results with Clearpath Jackal.} We conducted navigation deployments on different embodied platforms to show its applicability and robustness.}
\label{fig::real_2}
\end{figure}


\begin{table*}[t]
\small
\centering
\setlength{\tabcolsep}{4.2pt}
\begin{tabular}{lc|ccccc}
\toprule 
\textbf{Methods} & \textbf{With Map} & \textbf{1-time Run} & \textbf{10 Runs} & \textbf{20 Runs} & \textbf{30 Runs} & \textbf{60 Runs} \\
\midrule
ZSON\cite{majumdar2022zson}     & X      & 0.102 & 0.116 & 0.125 & 0.119 & 0.121 \\
PSL\cite{psl}      & X      & 0.224 & 0.215 & 0.197 & 0.202 & 0.199 \\
L3MVN\cite{yu2023l3mvn}    & $\surd$ & 0.186 & 0.217 & 0.252 & 0.281 & 0.358 \\
ESC\cite{zhou2023esc}      & $\surd$ & 0.249 & 0.286 & 0.337 & 0.351 & 0.389 \\
LOG-Nav & $\surd$ & 0.083 & 0.417 & 0.533 & 0.592 & 0.656 \\
\bottomrule 
\end{tabular}
\caption{\textbf{SPL comparison with ZSON approaches across different numbers of runs.}}
\label{additional_exp}
\end{table*}

\subsection{Simulation Experiments}

\textbf{Dataset.} MP3D~\citep{ramakrishnan2021hm3d} is a high-res photorealistic 3D reconstruction dataset of real-world scenes. We validate our approach on over 20 multi-room indoor scenarios. The dataset is formatted in a Habitat manner and defines six object categories: chairs, couches, potted plants, beds, toilets, and TVs. Notably, our approach considers layout-level information, enabling refined instance navigation experiments by differentiating objects in separate rooms, such as a chair in the living room versus one in the dining room, as distinct instances.

\textbf{Setups.} We employ the Habitat platform as our 3D indoor simulator. Observations resolution from the RGB-D camera is set to $480\times 640$, as used by~\citep{yu2023l3mvn}. Pose data is provided by an odometry sensor. The automatic exploration process is implemented using a simple policy approach, where the robot drives continuously along the left frontier while keeping the camera oriented to the right. Frames frequency is set to $15 Hz$, following~\citep{yang2023iplanner}. After traversing, the camera is reset to face forward. Our implementation is based on publicly available code examples from the project repository~\citep{puig2023habitat3}.

\textbf{Metrics.} We evaluate our approach using Success Rate (SR) and SR weighted by inverse Path Length (SPL), following established metrics in target navigation research~\citep{anderson2018evaluation}. Additionally, we provide instance-level navigation experiments, leveraging our approach's ability to recognize objects in different rooms as separate instances, capitalizing on the considered layout information.

\textbf{Comparison Results and Discussions.} As shown in Tab.\ref{exp_sim} and Fig.\ref{fig::sim_experiment}, our Log-Nav outperforms existing methods in both SR and SPL. The compared methods include L3MVN~\citep{yu2023l3mvn}, ESC~\citep{zhou2023esc}, and VLFM~\citep{yokoyama2024vlfm}, which employ LFM-powered mapping of separately frontier, probabilistic commonsense constraints, and visual-language similarity scoring. HOV-SG~\citep{werby23hovsg} employs global topological map, however, the static topological way-points extracted from occupancy cannot ensure efficient path planning or manage unexpected local scene changes which cannot ensure efficiency and robustness. This underscores the advantages of our hierarchical planning strategy, ensuring both effectiveness and efficiency by incorporating refined local waypoints in a cognitive-like global layout context~\citep{zeng2022theory}. Moreover, when comparing performances across object and instance navigation tasks, our method excels by setting the final target destination to the appropriate region based on related topo-vertices, overcoming limitations of previous methods that ignore layout.

Further, we compare with zero-shot-object-navigation (ZSON) methods repeatedly within the same workspace to complete object navigation tasks, shown in Tab. \ref{additional_exp}. All methods shared identical object goal sequences, and we compared the average SPL across different numbers of runs. Our 'explore-then-plan' approach constructs a complete map only once before performing planning cycles, whereas other 'mapping-during-navigation' ZSON methods accumulate mapping efforts over repeated executions. Results demonstrate that our mapping cost becomes amortized across planning iterations, particularly beneficial in scenarios with frequent deployments and extensive environments. By leveraging hierarchical environmental priors, our method outperforms incremental mapping approaches.

\subsection{Real-world Deployment}

Our real-world robot platforms include two types: one comprises a SLAMTEC mobile base, a Franka Panda arm, and a RealSense D435 camera mounted on the end. The camera is calibrated to the arm's base coordinates using the easy-hand-eye package. RGB-D images are recorded at $15 Hz$, matching the resolution of $480\times 640$ as in~\citep{yu2023l3mvn, yang2023iplanner}. The other one is a Clearpath robot with the same camera setting. The environment consists of a large-scale, multi-room indoor scene covering approximately $225 m^2$. It includes a small kitchen tabletop, office area, meeting room, and a hall. As demonstrated in Fig.\ref{fig::real_experiment}, the robot is tasked with navigating to different object instances through long-term paths.

To validate the robustness, after the robot explores and constructs the scene representation, we introduce obstacles and make minor adjustments to the scenario. This setup allows us to assess the navigation success rate under dynamic conditions. As shown in Fig.\ref{fig::real_experiment}, the robot successfully bypasses the obstacles or plans new paths. For further evaluation shown in Tab.\ref{exp_real}, we evaluate the path length, time cost to the destination, success rate, and number of global/local re-planning with each experiment setup 10 times. Additional results and task videos are shown in the website.

\subsection{Ablations}

\begin{figure}[t]
\centering
\includegraphics[width=\linewidth]{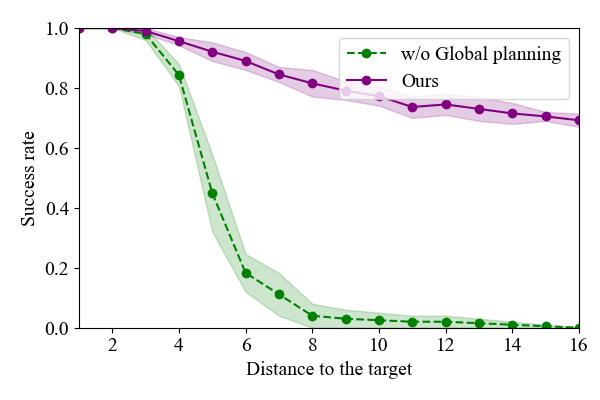}
\caption{\textbf{Ablations on Global Planning strategy.} Results show how SPL decreases with distance at different settings. "w/o" means "without".}
\label{fig:ablation2}
\end{figure}

\begin{table}[t]
\small
\centering
\setlength{\tabcolsep}{4.2pt}
\begin{tabular}{cccccc}
\toprule 
\multirow{2}{*}{Obstacles} & w/o Local & Planning&  & LOG-Nav & (Ours) \\ \cline{2-3} \cline{5-6}
                      & Cost (s) & SR & & Cost (s) & SR \\ 
\midrule
0              & 102 & 10/10 & & 96 & 10/10 \\
1              & 111 & 7/10 & & 103 & 10/10 \\
2              & 122 & 3/10 & & 112 & 8/10 \\
3              & 134 & 1/10 & & 116 & 6/10 \\
\bottomrule 
\end{tabular}
\caption{\textbf{Ablations on Local Planning strategy.} We evaluate the navigation time cost (only counting the successful navigation) and SR under different numbers of obstacles set in the way after exploration. Each navigation task is evaluated 10 times with similar settings except for the obstacles.}
\label{ablation3}
\end{table}


\textbf{Global Planning.} The global planning process incorporates topological information, from the local IL planning. To assess the impact of global planning, we disable it and directly project the long-term final destination's 3D position into the current observation frame, relying solely on the local IL planning. As illustrated in Fig.\ref{fig:ablation2}, the SR significantly decreases as the distance to targets increases, indicating local IL planning alone is insufficient for long-term planning.

\textbf{Local Planning.} This ablation is conducted in real-world setting. Without using local planning, we directly call the point navigation API of the mobile robot base to reach the globally planned way-point at each segment. The API allows robots to either successfully navigate to the target or attempt to circumvent obstacles as they arise. The time taken by the robot to reach the destination is recorded and analyzed. As shown in Tab.\ref{ablation3}, the SR and efficiency drop rapidly when obstacle numbers grow without local planning.


%% file: sections/limitations.tex
\section{Limitations}\label{limitations}

This paper introduces LOG-Nav, an efficient layout-aware object-goal navigation approach for multi-room indoor environments, hierarchically planning on both global and local levels. 
Our approach makes progress in addressing globally and locally efficient long-term path planning in complex indoor scenes and ease of deployment without human aid, complex rewards, or costly training.
Despite these advancements, our approach currently has two limitations: 1) While local mapping is implemented through imperative learning, it does not fully leverage the potential of the local scene memory for refined obstacle avoidance and improved efficiency. 2) The detected scene changes are not dynamically integrated into the scene representation and topological map. Addressing these limitations by implementing real-time scene updates remains a priority for future work.